\documentclass[11pt,a4paper]{article}
\usepackage{jheppub}
\usepackage[T1]{fontenc}
\usepackage{lmodern}
\usepackage{microtype}
\usepackage{booktabs}
\usepackage{array}
\usepackage{textcomp}
\usepackage{multirow}
\usepackage{siunitx}
\usepackage{listings}
\usepackage{tikz}
\usetikzlibrary{positioning,arrows.meta,shapes.geometric,calc,fit,backgrounds}
\usepackage{xurl}

\newcommand{\NumSubtasks}{11}
\newcommand{\TaskXcorrDelay}{\code{xcorr\_delay}}
\newcommand{\TaskXcorrFine}{\code{xcorr\_fine}}

\newcommand{\TaskFFTBinSnr}{\code{fft\_bin\_snr}}

\newcommand{\NumBuiltins}{445}
\newcommand{\MaxElems}{20\,million}
\newcommand{\ScoreTimeout}{120}
\newcommand{\MaxTokensLimit}{5000}
\newcommand{\NumExamples}{66}

\newcommand{\HeldOutSeen}{\code{fft\_noise\_power} variant 7}
\newcommand{\NumThoughtExamples}{64}
\newcommand{\NumNaturalExamples}{2}
\newcommand{\NumHandFixed}{25}
\newcommand{\NumCodeFixed}{10}
\newcommand{\NumTextFixed}{15}
\newcommand{\ThoughtWordsMin}{228}
\newcommand{\ThoughtWordsMax}{436}
\newcommand{\ThoughtWordsMean}{310}
\newcommand{\NumFormulaOnly}{3}
\newcommand{\NumRulesStepThree}{17}
\newcommand{\NumKeyPointsSameVariant}{9}
\newcommand{\NumFirstCodeAccepted}{43}
\newcommand{\MaxCodeAttempts}{3}
\newcommand{\NumRuleFacts}{22}
\newcommand{\FirstRuns}{53}
\newcommand{\FirstPass}{16}
\newcommand{\FirstZeroSubtasks}{4}
\newcommand{\LoraRank}{16}
\newcommand{\LoraAlpha}{16}
\newcommand{\Epochs}{3}
\newcommand{\LearningRate}{10^{-4}}

\newcommand{\EfourTpsBase}{15.9}
\newcommand{\EfourTpsAdapter}{13.0}
\newcommand{\EfourTpsDropPct}{19}
\newcommand{\SpeedTokens}{1024}
\newcommand{\EtwoTpsBase}{30.7}
\newcommand{\EtwoTpsAdapter}{25.4}
\newcommand{\EtwoTpsDropPct}{17}
\newcommand{\EfourTpsAdapterNew}{12.9}
\newcommand{\EfourTpsAdapterOld}{13.2}
\newcommand{\EtwoBaseSecPerRun}{50}

\newcommand{\EtwoLoraSecPerRun}{42}

\newcommand{\EfourBaseSecPerRun}{121}

\newcommand{\EfourQloraSecPerRun}{106}

\newcommand{\MaxTokensUsed}{3063}

\newcommand{\SeedFirst}{5000}
\newcommand{\SeedLast}{5002}
\newcommand{\NumSeeds}{3}
\newcommand{\EtwoBasePass}{18}
\newcommand{\EtwoBaseScore}{46.5}
\newcommand{\EtwoBaseTokens}{1769}
\newcommand{\EtwoLoraPass}{37}
\newcommand{\EtwoLoraScore}{71.3}
\newcommand{\EtwoLoraTokens}{1286}
\newcommand{\EtwoQloraPass}{37}
\newcommand{\EtwoQloraScore}{69.9}

\newcommand{\EfourBasePass}{32}
\newcommand{\EfourBaseScore}{68.7}
\newcommand{\EfourBaseTokens}{1713}
\newcommand{\EfourQloraPass}{47}
\newcommand{\EfourQloraScore}{84.7}
\newcommand{\EfourQloraTokens}{1283}

\newcommand{\NumRunsUnseen}{63}
\newcommand{\EtwoBaseNightlyRuns}{16}
\newcommand{\EtwoBaseNightlyPass}{1}
\newcommand{\EtwoBaseOldRunsFour}{8}
\newcommand{\EtwoBaseOldPassFour}{0}
\newcommand{\CheckRunTask}{\code{xcorr\_delay} variant 6, seed 5000}
\newcommand{\CheckRunNightly}{100}

\newcommand{\EtwoBaseErrConvention}{17}
\newcommand{\EtwoBaseErrUnits}{15}
\newcommand{\EtwoBaseErrShape}{9}

\newcommand{\EtwoBaseErrMatched}{4}

\newcommand{\EtwoBaseErrDelayed}{6}
\newcommand{\EtwoBaseErrNoise}{11}
\newcommand{\EtwoBaseErrFFTgain}{11}
\newcommand{\EtwoBaseErrWaveform}{13}

\newcommand{\EtwoLoraErrConvention}{6}
\newcommand{\EtwoLoraErrUnits}{9}
\newcommand{\EtwoLoraErrShape}{1}

\newcommand{\EtwoLoraErrMatched}{0}

\newcommand{\EtwoLoraErrDelayed}{6}
\newcommand{\EtwoLoraErrNoise}{7}
\newcommand{\EtwoLoraErrFFTgain}{3}
\newcommand{\EtwoLoraErrWaveform}{8}

\newcommand{\EtwoQloraErrConvention}{5}
\newcommand{\EtwoQloraErrUnits}{7}
\newcommand{\EtwoQloraErrShape}{3}

\newcommand{\EtwoQloraErrMatched}{1}

\newcommand{\EtwoQloraErrDelayed}{4}
\newcommand{\EtwoQloraErrNoise}{7}
\newcommand{\EtwoQloraErrFFTgain}{4}
\newcommand{\EtwoQloraErrWaveform}{10}

\newcommand{\EfourBaseErrConvention}{18}
\newcommand{\EfourBaseErrUnits}{4}

\newcommand{\EfourBaseErrNoise}{7}
\newcommand{\EfourBaseErrFFTgain}{5}

\newcommand{\EfourQloraErrConvention}{2}
\newcommand{\EfourQloraErrUnits}{0}

\newcommand{\EfourQloraErrNoise}{6}
\newcommand{\EfourQloraErrFFTgain}{5}

\newcommand{\NumFailsRead}{159}
\newcommand{\EfourQloraXinceFails}{5}
\newcommand{\EfourQloraXinceNoiseOnly}{3}
\newcommand{\EtwoChirpFails}{25}
\newcommand{\EtwoChirpWaveform}{22}
\newcommand{\NumErrClasses}{12}
\newcommand{\EtwoBasePassXcorrGecikme}{0}

\newcommand{\EtwoBasePassKonvGecikme}{0}
\newcommand{\EtwoBasePassFftGurultuGucu}{1}

\newcommand{\EtwoBasePassFmcwFrekans}{1}
\newcommand{\EtwoBasePassLfmTek}{3}

\newcommand{\EtwoLoraPassXcorrGecikme}{2}

\newcommand{\EtwoLoraPassKonvGecikme}{3}
\newcommand{\EtwoLoraPassFftGurultuGucu}{4}

\newcommand{\EtwoLoraPassFmcwFrekans}{1}
\newcommand{\EtwoLoraPassLfmTek}{3}

\newcommand{\EtwoQloraPassXcorrGecikme}{5}

\newcommand{\EtwoQloraPassKonvGecikme}{2}
\newcommand{\EtwoQloraPassFftGurultuGucu}{5}

\newcommand{\EtwoQloraPassFmcwFrekans}{1}
\newcommand{\EtwoQloraPassLfmTek}{2}

\newcommand{\EfourBasePassXcorrGecikme}{0}

\newcommand{\EfourBasePassKonvGecikme}{1}
\newcommand{\EfourBasePassFftGurultuGucu}{4}

\newcommand{\EfourBasePassLfmTek}{3}

\newcommand{\EfourQloraPassXcorrGecikme}{4}

\newcommand{\EfourQloraPassKonvGecikme}{4}
\newcommand{\EfourQloraPassFftGurultuGucu}{3}

\newcommand{\EfourQloraPassLfmTek}{6}

\newcommand{\EtwoBaseEtwoLoraOnlyB}{22}
\newcommand{\EtwoBaseEtwoLoraOnlyA}{3}
\newcommand{\EtwoBaseEtwoLoraP}{0.00016}
\newcommand{\EtwoBaseEtwoQloraOnlyB}{25}
\newcommand{\EtwoBaseEtwoQloraOnlyA}{6}
\newcommand{\EtwoBaseEtwoQloraP}{0.00088}
\newcommand{\EtwoLoraEtwoQloraOnlyB}{11}
\newcommand{\EtwoLoraEtwoQloraOnlyA}{11}

\newcommand{\EfourBaseEfourQloraOnlyB}{21}
\newcommand{\EfourBaseEfourQloraOnlyA}{6}
\newcommand{\EfourBaseEfourQloraP}{0.0059}

\newcommand{\EfourBaseEtwoLoraOnlyB}{13}
\newcommand{\EfourBaseEtwoLoraOnlyA}{8}
\newcommand{\EfourBaseEtwoLoraP}{0.38}

\definecolor{cbase}{HTML}{77776F}   % base model (no adapter)
\definecolor{clora}{HTML}{2A78D6}   % LoRA adapter
\definecolor{cqlora}{HTML}{EB6834}  % QLoRA adapter
\definecolor{cgreen}{HTML}{1BAF7A}
\definecolor{cink}{HTML}{3D3D3A}
\definecolor{cshade}{HTML}{E9EEF6}

\tikzset{
  blk/.style={draw=cink, rounded corners=2pt, align=center, minimum height=8mm,
              inner sep=3pt, font=\small, fill=white},
  nl/.style={blk, draw=cqlora, line width=0.9pt},
  io/.style={blk, fill=cshade},
  arr/.style={-{Stealth[length=2mm]}, draw=cink},
  note/.style={font=\footnotesize, text=cink, align=left},
}

\newcommand{\figref}[1]{figure~\ref{#1}}
\newcommand{\Figref}[1]{Figure~\ref{#1}}
\newcommand{\tabref}[1]{table~\ref{#1}}
\newcommand{\Tabref}[1]{Table~\ref{#1}}
\newcommand{\secref}[1]{section~\ref{#1}}
\newcommand{\Secref}[1]{Section~\ref{#1}}
\newcommand{\appref}[1]{appendix~\ref{#1}}
\newcommand{\Appref}[1]{Appendix~\ref{#1}}

\newcommand{\code}[1]{\texttt{#1}}

\title{Fine-Tuning Gemma~4 Edge Models for MATLAB Signal-Processing Code with SFT and QLoRA on a Laptop iGPU}
\author{Mehmet Altan Toks\"oz, PhD}
\affiliation{Independent researcher, G\"olba\c{s}\i, Ankara, T\"urkiye}
\emailAdd{matoksoz@gmail.com}
\subheader{\today}
\abstract{Small language models that run on a laptop are attractive for writing signal-processing code
offline, but they make systematic errors on digital signal processing (DSP) conventions.
We apply supervised fine-tuning (SFT) to the Gemma~4 edge models E2B and E4B with 66 checked examples
covering 11 signal-processing MATLAB subtasks, using low-rank adaptation (LoRA) and quantized LoRA
(QLoRA) on the integrated GPU (iGPU) of a laptop, and deploy the 8-bit integer (int8) models with the
adapters through OpenVINO.
On held-out variants scored by deterministic checkers, the pass rate rises from
\EtwoBasePass/66 to \EtwoLoraPass/66 (LoRA) and \EtwoQloraPass/66 (QLoRA) for E2B and from
\EfourBasePass/66 to \EfourQloraPass/66 for E4B (QLoRA).
Reading all \NumFailsRead{} failing programs shows that the adapters remove most errors of convention,
such as the sign of a lag, units and array shapes, but few errors in building signals and noise.
Applied at run time, the adapter costs \EtwoTpsDropPct--\EfourTpsDropPct\% of the generation speed.
}

\begin{document}\maketitle

\section{Introduction}\label{sec:intro}

Signal-processing engineers write much of their code in
MATLAB: they synthesise pulses and chirps, correlate channels, filter, take fast Fourier transforms (FFTs) and set detection
thresholds.
Language models can draft such code, and small models that run on a laptop can do so without
sending anything off the machine, which matters where the work cannot leave it.
The Gemma~4 edge models E2B and E4B \cite{gemma4} are made for this kind of device.
On our tasks, however, they often get the conventions of the field wrong: they read the delay with
the wrong sign from \code{xcorr}, forget the gain $N$ of an unscaled FFT, write a chirp rate in the
wrong unit or mix row and column vectors.
Such code runs and prints a plausible number, which makes the errors hard to notice.

Low-rank adaptation (LoRA) \cite{hu2022lora} and its quantized variant QLoRA \cite{dettmers2023qlora}
make supervised fine-tuning (SFT) of such models possible with little memory.
We ask how much a small set of checked examples, trained on the integrated GPU (iGPU) of the same laptop,
improves the code these models write, and which errors it removes.

Answering this needs three things.
First, tasks whose answers can be checked automatically.
Following the practice of judging generated code by running it \cite{chen2021codex}, we wrote
\NumSubtasks{} parametric subtasks from signal-processing work, run every answer in an emulator of a
MATLAB subset, and score it with checkers that also reject values copied from the prompt or computed
from a formula instead of measured.
Earlier work compared commercial and open models on generating MATLAB code for automated-driving
scenarios \cite{yang2025matlab}; here the models are small enough to run on a laptop and the domain
is digital signal processing (DSP).
Second, training data.
Instead of collecting the model's own correct answers, which were too rare, we let the small model
write the reasoning for an expert's solution and then the answer itself, close to the
rationalization step of the Self-Taught Reasoner (STaR) \cite{zelikman2022star}, and kept only
answers that pass the checkers.
Third, a way to train and deploy on a laptop: the Parameter-Efficient Fine-Tuning (PEFT) library
\cite{peft} on the integrated GPU for training, and OpenVINO GenAI \cite{openvinogenai} with 8-bit
integer (int8) weights for inference.

Our contributions are:
\begin{itemize}
\item \NumSubtasks{} parametric MATLAB signal-processing subtasks with deterministic
checkers; the prompts of the held-out variants are listed in \appref{app:prompts};
\item a procedure in which the small model itself writes \NumExamples{} training examples from expert
solutions, with automatic filters and a reading of every example;
\item LoRA and QLoRA adapters for E2B and E4B trained on a laptop GPU in under an hour each and run
with int8 OpenVINO models;
\item an evaluation on held-out variants in which the adapters double the pass count of E2B and raise
that of E4B from \EfourBasePass{} to \EfourQloraPass{} of 66, and an analysis of all
\NumFailsRead{} failing programs that shows which errors the adapters remove and which they do not.
\end{itemize}

\Secref{sec:tasks} describes the subtasks, the emulator and the checkers, \secref{sec:data} the
training data and \secref{sec:training} the training and deployment.
\Secref{sec:results} gives the results, \secref{sec:errors} the error analysis and
\secref{sec:limitations} the limitations.

\section{Tasks and evaluation}\label{sec:tasks}

\subsection{Subtasks}\label{sec:subtasks}

The tasks come from signal processing: delay estimation between
two receive channels, matched filtering, FFT-based detection, frequency-modulated continuous-wave (FMCW) and linear frequency-modulated
(LFM) waveform generation, and synthetic pulse descriptor words (PDWs), which list each pulse with
its time of arrival (TOA) and other parameters.
In a first measurement, E2B failed every run of most of these complete tasks, so we split them
into the steps at which the models failed.
This gives the \NumSubtasks{} subtasks of \tabref{tab:subtasks}.
Each subtask asks for a short MATLAB script that builds a signal, processes it and prints a few
numbers.

Every subtask is a parametric template.
A variant is drawn from a random generator seeded with the subtask name and the variant index, so
every variant is reproducible.
The generator draws sampling rates, lengths, delays, bandwidths, amplitudes and signal-to-noise ratios (SNRs), writes the
prompt in English and computes the expected values with NumPy and SciPy.
A variant is drawn again when an expected value equals a number in the prompt or a frequent
constant (2, 10, 20, 100): such a value could be copied instead of computed.
Variants 0--5 of each subtask are used to make training data (\secref{sec:data}); variants 6 and 7
are held out for evaluation.
The held-out variants come from the same distribution as the training variants, so the evaluation
measures generalisation to new parameters of known subtasks, not to new subtasks.
Each template also has a reference solution, which scores 100 on both held-out variants of every
subtask.
\Appref{app:prompts} lists the prompt of variant 6 of every subtask with its expected values.

\begin{table}[t]
\centering\small
\begin{tabular}{@{}r l >{\raggedright\arraybackslash}p{0.34\textwidth} >{\raggedright\arraybackslash}p{0.34\textwidth}@{}}
\toprule
& Label & Subtask & Checks \\
\midrule
1 & \code{xcorr\_delay} & integer delay between two LFM channels from the \code{xcorr} peak & integer delay (from \code{xcorr}); channel 2 = delayed channel 1$^w$ \\
2 & \code{xcorr\_fine} & fractional delay in noise: \code{xcorr} peak refined by a parabola & refined delay ($\pm 0.2$ samples)$^r$; channel 2 = delayed channel 1$^w$; noise level of both channels$^w$ \\
3 & \code{noise\_power} & complex tone plus noise at a given SNR; measured powers & noise power$^r$; variance of the real part$^r$; power of $x$$^r$; $x$ = tone + noise$^w$ \\
4 & \code{matched\_output} & matched filter of an LFM pulse with \code{conv}: length, peak index, peak & length; peak index; peak magnitude \\
5 & \code{conv\_delay} & delay of a pulse in a receive window from the matched-filter peak & peak index; delay from the peak index \\
6 & \code{fft\_noise\_power} & mean $|X|^2$ of complex noise after an unscaled FFT, measured and theoretical & measured mean$^r$; theoretical mean \\
7 & \code{exp\_threshold} & threshold on $|X(k)|^2$ for a false-alarm probability (exponential law) & threshold on $|X|^2$; threshold on $|X|$ \\
8 & \code{fft\_bin\_snr} & tone in noise: peak bin, measured and theoretical SNR after the FFT & peak bin (1-based); measured SNR$^r$; theoretical SNR \\
9 & \code{fmcw\_freq} & instantaneous-frequency vector of a sawtooth or triangular FMCW signal & length; minimum; maximum; value at a given time \\
10 & \code{lfm\_single} & one complex baseband linear chirp stored in a vector & instantaneous frequency of the vector$^w$ \\
11 & \code{pdw\_single} & pulse descriptor words (PDW matrix) of one emitter with constant pulse repetition interval (PRI) & number of PDWs; last TOA; PDW matrix$^w$; rows sorted by TOA$^w$ \\
\bottomrule
\end{tabular}

\caption{The \NumSubtasks{} subtasks and the checks that score them.
$^r$: noise-dependent value, accepted only if it changes when the code runs again with another seed.
$^w$: workspace check on the arrays left by the code rather than on the printed output.
Every subtask also scores the error-free run of the code.}
\label{tab:subtasks}
\end{table}

\subsection{MATLAB emulator}\label{sec:emulator}

We run the generated code in our own MATLAB emulator, since no MATLAB installation was available.
The emulator parses a subset of the MATLAB language into a syntax tree and evaluates it on NumPy
arrays; it does not translate MATLAB to Python and never calls \code{exec}.
It provides \NumBuiltins{} built-in functions, among them \code{xcorr}, \code{conv}, \code{fft},
\code{randn}, \code{repmat}, \code{interp1} and \code{polyfit}.
Code cannot call \code{system} or \code{eval}, cannot read or write files and can create no array with
more than \MaxElems{} elements.
Error messages start with MATLAB's own wording, so a model that knows MATLAB errors sees familiar
text.
N-dimensional arrays, struct arrays, integer types, \code{global} and \code{classdef} are not supported.
The emulator can differ from MATLAB in corner cases; we return to this in \secref{sec:limitations}.

\subsection{Checkers and score}\label{sec:checkers}

The code is taken from the \code{matlab} blocks of the answer and run as written: there is no
automatic repair, and the emulator's own hints are switched off.
A run stops after \ScoreTimeout{}~s.
The checker parses every number of the printed output and looks for each expected value within a
tolerance.
Several rules make sure that a value is computed by the code rather than copied or derived from the
prompt.
\begin{itemize}
\item Numbers written as literals in the code (comments excluded) are not accepted as results.
\item A noise-dependent value (marked $^r$) is accepted only if it changes when the code runs a
second time with another seed and with its own \code{rng} call disabled.
A number printed identically twice was computed from a formula, not measured.
\item A value that must come from a peak index (the integer delay and the matched-filter peak) is
checked by a second run in which the output of \code{xcorr} or \code{conv} is shifted by one
sample.
The printed value must move with it.
A delay printed from \code{tau*fs} does not move.
\item Some values count only if the code called the function the prompt names (\code{xcorr} or
\code{conv}).
\item Workspace checks (marked $^w$) inspect the arrays the code leaves behind.
They check that channel 2 is channel 1 delayed by $\tau$, that the noise outside the pulse has the
requested level in both channels, that $x$ is the tone plus noise, that the instantaneous frequency
of the stored chirp or FMCW vector follows the requested profile, and that the PDW matrix holds the
right rows.
\end{itemize}
The score of a run is 20 points for running without error plus 80 points shared among the checks of
the subtask in proportion to their weights.
All weights are 1, except the FMCW signal check (4) and the PDW matrix check (3).
A run \emph{passes} when it scores 100, that is, when the code runs and every check passes.

\subsection{Protocol}\label{sec:protocol}

Each model answers each held-out variant once per seed, with seeds \SeedFirst--\SeedLast{}.
This gives $\NumSubtasks \times 2 \times \NumSeeds = 66$ runs per model and adapter.
The answer is one-shot: one system message
(``You are an assistant that writes MATLAB code. Give the code in a \texttt{\textasciigrave\textasciigrave\textasciigrave matlab} block.''),
the English prompt, no tools and no second attempt.
Thinking is on, as the Gemma~4 chat template allows.
The answer may use up to \MaxTokensLimit{} tokens; the longest answer had \MaxTokensUsed{} tokens and
no run hit the limit.
The models run as int8 OpenVINO models on the laptop's integrated GPU behind an OpenAI-compatible
server (\secref{sec:training}).
E2B samples with temperature 0.9 and E4B with temperature 0.8, both with top-$p$ 0.95 and top-$k$ 50.
The E2B setting comes from an earlier search over sampling parameters on other MATLAB tasks; the E4B
setting is the one we used with E4B before this work.

\section{Training data}\label{sec:data}

The training set has one example for each training variant: \NumExamples{} examples, six per subtask.
An example is a conversation with the system message and the prompt of \secref{sec:protocol},
followed by the model's thinking and its answer with the MATLAB code, so training and evaluation
use the same format.
In a first measurement on training variants, E2B passed \FirstPass{} of \FirstRuns{} runs and none
of the runs of \FirstZeroSubtasks{} subtasks, so rejection sampling of its own answers would have
left several subtasks without data and others with very few examples.
We therefore made the examples from expert reference solutions, but had E2B write all the text
that enters the training set: the thinking and the answer.
No larger model writes training text; a larger model wrote only the list of facts that guided
some answers in step~3 (see below).
\Figref{fig:pipeline} shows the three steps; \NumThoughtExamples{} examples come from this
pipeline and \NumNaturalExamples{} are answers that E2B wrote on its own and that passed.

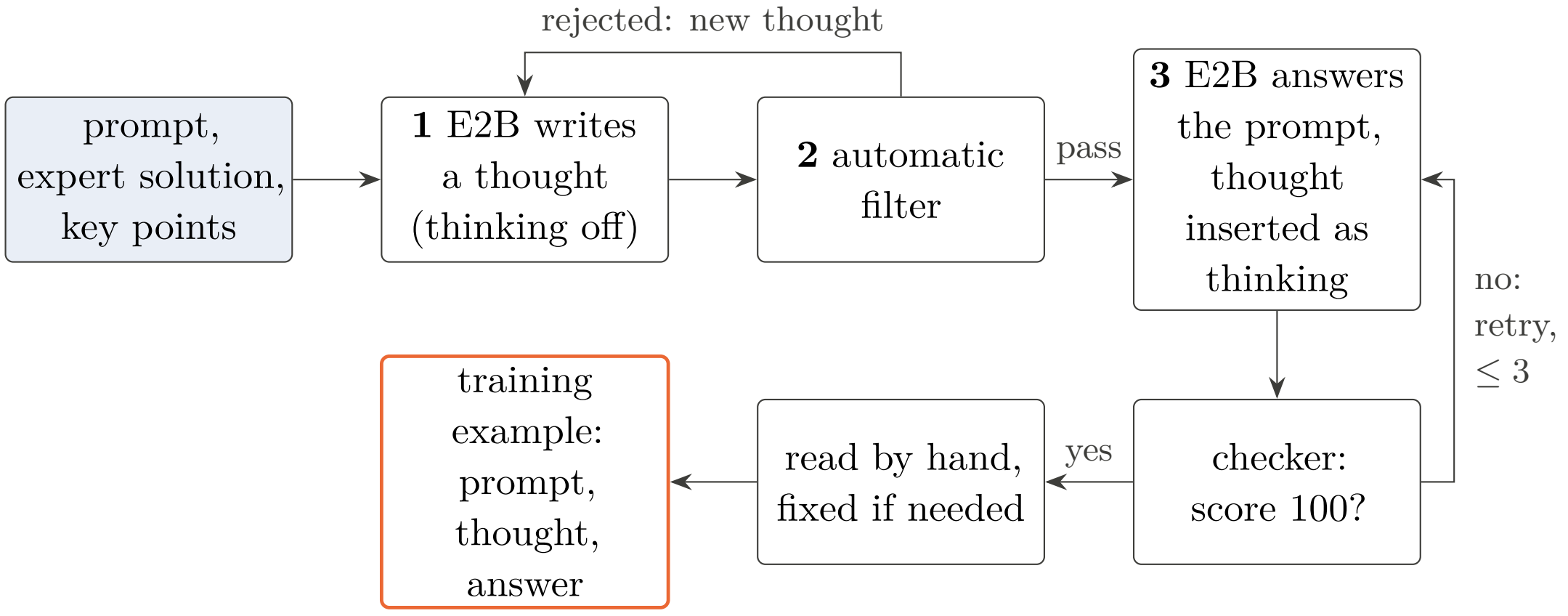
\begin{figure}[t]
\centering
\begin{tikzpicture}[node distance=8mm and 8mm,
                    b/.style={blk, text width=24mm, minimum height=15mm}]
\node[io, text width=24mm, minimum height=15mm] (in) {prompt,\\ expert solution,\\ key points};
\node[b, right=of in] (s1) {\textbf{1}~E2B writes a thought (thinking off)};
\node[b, right=of s1] (s2) {\textbf{2}~automatic\\ filter};
\node[b, right=of s2] (s3) {\textbf{3}~E2B answers the prompt, thought inserted as thinking};
\node[b, below=of s3] (ck) {checker:\\ score 100?};
\node[b] (hr) at (s2 |- ck) {read by hand, fixed if needed};
\node[b, nl] (ex) at (s1 |- ck) {training example: prompt, thought, answer};
\draw[arr] (in) -- (s1);
\draw[arr] (s1) -- (s2);
\draw[arr] (s2) -- node[note, above] {pass} (s3);
\draw[arr] (s3) -- (ck);
\draw[arr] (ck) -- node[note, above] {yes} (hr);
\draw[arr] (hr) -- (ex);
\draw[arr] (s2.north) -- ++(0,4mm) -| node[note, above, pos=0.25] {rejected: new thought} (s1.north);
\draw[arr] (ck.east) -- ++(3mm,0) |- (s3.east);
\node[note, anchor=west] at ($(ck.east)!0.5!(s3.east) + (3.5mm,0)$) {no:\\ retry,\\ $\le 3$};
\end{tikzpicture}
\caption{How the training examples were made.
Step 1 gives E2B the prompt with the expert solution and key points; step 3 gives it only the
system message and the prompt, with the accepted thought inserted as its own thinking, so the
answer is written as at evaluation time.
Only the thought of step 1 and the answer of step 3 enter the training example.}
\label{fig:pipeline}
\end{figure}

\paragraph{Step 1: thought.}
E2B receives, with thinking off, the prompt of the variant, an expert's reference solution of that
variant and the key points of the subtask: a short paragraph that states its decisive steps,
such as which lag of \code{xcorr} is the delay.
The key points are written for one variant of each subtask; for \NumKeyPointsSameVariant{} of the
\NumThoughtExamples{} accepted thoughts they belong to the same variant, otherwise they are
introduced as written for a similar task with other numbers.
E2B is asked to write the thinking an assistant would go through before writing code, in its own
style and in at most 300 words.
The instruction forbids mentioning the solution, asks to derive every number from the task and
never to state a noise-dependent result as a value, to state each decision exactly, and to reason
once without correcting itself.
For \NumFormulaOnly{} subtasks (\TaskFFTBinSnr, \TaskXcorrDelay, \TaskXcorrFine)
the instruction also forbids arithmetic: every derived value stays a formula or a MATLAB
expression, such as \code{tau*fs} or \code{lags(idx)}, for the code to compute, and every value with a
unit is restated in seconds, hertz or Hz/s.
In earlier attempts E2B computed bins and delays by hand and got them wrong, and those numbers then
went into the code.

\paragraph{Step 2: filter.}
A thought is rejected when it is shorter than 150 or longer than 900 words, contains a code block,
refers to the solution or the key points, corrects itself (\emph{wait}, \emph{hmm}), uses vague
phrases (\emph{depending on}), or writes \code{randn(N)}, which gives an $N\times N$ matrix in MATLAB.
Every number in the thought must be traceable: it must be a number of the prompt, a value of the
reference solution that does not change with the noise seed, or one step away from these (half,
double, inverse, square root, dB, a product or quotient of two such numbers, a unit scale).
A number that matches only a noise-dependent value of the solution is a measurement the thought
cannot know and is rejected.
For the formula-only subtasks only the numbers of the prompt and simple one-step forms are allowed,
the chirp rate must be restated in Hz/s, and the thought must say that the delay is read from the
\code{lags} output of \code{xcorr}.
For the other subtasks every integer the task asks to print (an index, length or count) must appear
in the thought as a number.

\paragraph{Step 3: answer.}
The accepted thought is inserted as E2B's own thinking in a request that contains only the system
message and the prompt, and E2B continues with the answer.
The answer is scored by the checkers of \secref{sec:checkers}, and the first answer that scores 100
is accepted; this happened at the first attempt for \NumFirstCodeAccepted{} of the
\NumThoughtExamples{} examples and never needed more than \MaxCodeAttempts{} attempts.
For \NumRulesStepThree{} examples this step used a longer system message with \NumRuleFacts{} numbered
facts about sampling, FFT scaling, correlation and noise, written by Claude Opus 5.5.
These facts served only to obtain a correct answer: the training example keeps the plain system
message, answers that referred to the facts were rejected, and fact numbers were removed from the
answer text.

\paragraph{Reading and hand fixes.}
Every accepted example was read by hand, and \NumHandFixed{} of the \NumExamples{} were fixed.
In \NumCodeFixed{} examples the fix changed what the code computes or prints.
For example, five \TaskFFTBinSnr{} answers set the peak bin to zero and still averaged over all
$N$ bins, one \TaskXcorrFine{} answer added the same noise vector to both channels, and one
\TaskXcorrDelay{} answer built channel~2 by shifting channel~1 by a whole number of samples.
The checkers accepted these answers, because they test the printed values and some arrays, not
every step (\secref{sec:limitations}).
The other \NumTextFixed{} fixes changed only the thought, comments, variable names or unused lines,
for example a wrong intermediate number in a thought (\emph{0.057~MHz = 570{,}000~Hz}) or a chirp
rate called a carrier frequency.
After the fixes, every example still scores 100 with the current checkers.

\paragraph{The resulting set.}
The thoughts have \ThoughtWordsMin--\ThoughtWordsMax{} words, \ThoughtWordsMean{} on average.
For two subtasks the prompt of variant 3 repeated that of variant 0, and variant 8 was used instead.
While writing this paper we found that one held-out variant, \HeldOutSeen, has the same prompt as a
training example; \secref{sec:results} also reports the results without it.
The training set is not released; the prompts of the held-out variants are listed in
\appref{app:prompts}.

\section{Fine-tuning and deployment}\label{sec:training}

\paragraph{Hardware.}
Everything runs on one laptop with an Intel Core Ultra~5 226V processor, its integrated Arc 130V
GPU and 16~GB of memory (15.6~GB usable) shared between the CPU and the GPU.
Training uses the GPU through PyTorch~2.9.1 with its XPU back end (PyTorch's device type for Intel GPUs), Transformers~5.22 and PEFT~0.21;
QLoRA adds bitsandbytes~0.50.2.
Newer PyTorch builds gave wrong results with the installed GPU driver, so we kept 2.9.1.

\paragraph{Adapters.}
We train three adapters on the same \NumExamples{} examples (\tabref{tab:training}):
a LoRA adapter on the bfloat16 (bf16) weights of E2B, and QLoRA adapters on E2B and E4B, whose base
layers are quantized to the 4-bit NormalFloat (NF4) type with double quantization and computed in bf16.
E4B has only a QLoRA adapter, because its bf16 weights do not leave enough GPU memory for training.
The adapters have rank \LoraRank{} and $\alpha = \LoraAlpha$, no dropout, and sit on the query, key,
value and output projections and the three multilayer perceptron (MLP) projections of every language-model layer.
The loss is computed only on the tokens of the thought and the answer, with the prompt formatted by
the same chat template as at inference.
Training uses AdamW with learning rate $\LearningRate$ and no weight decay, one example per step,
gradient clipping at 1, gradient checkpointing and \Epochs{} epochs with the examples shuffled in
every epoch.
There was one training run per adapter; we did not tune these settings.

\begin{table}[t]
\centering\small
\begin{tabular}{lcrccr}
\toprule
Adapter & Base weights & Trainable & Loss per epoch & Time & Peak iGPU \\
\midrule
E2B LoRA & bf16 & 24.2\,M & 0.35 / 0.23 / 0.17 & 31~min & 6.9~GB \\
E2B QLoRA & NF4 & 24.2\,M & 0.38 / 0.25 / 0.19 & 53~min & 4.4~GB \\
E4B QLoRA & NF4 & 34.9\,M & 0.40 / 0.27 / 0.20 & 44~min & 6.3~GB \\
\bottomrule
\end{tabular}

\caption{The three adapters, trained on the same \NumExamples{} examples for \Epochs{} epochs.
Loss is the mean training loss of each epoch; time is the training time without loading; peak iGPU
is the largest memory allocated on the GPU during training.}
\label{tab:training}
\end{table}

\paragraph{Deployment.}
For inference we use the official int8 OpenVINO exports of the two models and OpenVINO GenAI on the
GPU, behind a small OpenAI-compatible server.
The adapter is loaded as saved by PEFT and applied at run time without merging it into the
weights, so the adapter trained on bf16 or NF4 weights runs on int8 weights.
The adapter evaluations used OpenVINO~2026.4.1.
The E4B base model failed on its first requests with an out-of-resources error of the GPU runtime in
that version, so the E4B base evaluation and the E2B base runs with seeds 5001--5002 on four subtasks used a
2026.5 nightly build; \secref{sec:results} discusses this difference.

\paragraph{Speed.}
Applying the adapter at run time costs speed.
On the E4B server, with the same three general prompts, greedy decoding and \SpeedTokens{} generated
tokens each, generation
ran at \EfourTpsBase{} tokens/s without the adapter and \EfourTpsAdapter{} tokens/s with it, a
\EfourTpsDropPct\% drop.
The same measurement on E2B gave \EtwoTpsBase{} tokens/s without and \EtwoTpsAdapter{} tokens/s with
the LoRA adapter, a \EtwoTpsDropPct\% drop.
For E4B with the adapter, the nightly build and 2026.4.1 gave the same speed
(\EfourTpsAdapterNew{} and \EfourTpsAdapterOld{} tokens/s).
Merging the adapter into the int8 weights would remove this cost; the run-time merge mode of
OpenVINO GenAI accepts only floating-point weights.
In the evaluations the adapters also shortened the answers, from \EfourBaseTokens{} to
\EfourQloraTokens{} tokens on average for E4B, so a run took \EfourQloraSecPerRun{}~s with the adapter
and \EfourBaseSecPerRun{}~s without (\EtwoLoraSecPerRun{}~s and \EtwoBaseSecPerRun{}~s for E2B with
and without LoRA).
These run times come from evaluations with different OpenVINO versions and are only indicative.

\section{Results}\label{sec:results}

\paragraph{Pass rates.}
\Tabref{tab:main} gives the results of the five evaluated models on the 66 held-out runs.
Both adapters double the pass count of E2B, from \EtwoBasePass{} to \EtwoLoraPass{} with LoRA and
\EtwoQloraPass{} with QLoRA, and the E4B adapter raises it from \EfourBasePass{} to \EfourQloraPass.
The mean score rises with the pass count, from \EtwoBaseScore{} to \EtwoLoraScore{} and
\EtwoQloraScore{} for E2B and from \EfourBaseScore{} to \EfourQloraScore{} for E4B.
The adapted models also write shorter answers: E4B with the adapter uses \EfourQloraTokens{} tokens per
answer instead of \EfourBaseTokens, E2B with LoRA \EtwoLoraTokens{} instead of \EtwoBaseTokens.
Without the held-out variant whose prompt is also a training prompt (\secref{sec:data}), the
counts change by at most two and the order of the models stays the same (column \emph{of
\NumRunsUnseen} in \tabref{tab:main}).

\begin{table}[t]
\centering\small
\begin{tabular}{lrrrrrl}
\toprule
& \multicolumn{2}{c}{Passed} & Mean & Mean & Time & \\
\cmidrule(lr){2-3}
Model & of 66 & of 63$^*$ & score & tokens & per run & OpenVINO \\
\midrule
E2B base & 18 & 17 & 46.5 & 1769 & 50~s & 2026.4.1 / 2026.5$^\dagger$ \\
E2B LoRA & \textbf{37} & \textbf{35} & \textbf{71.3} & 1286 & 42~s & 2026.4.1 \\
E2B QLoRA & \textbf{37} & \textbf{35} & 69.9 & 1546 & 48~s & 2026.4.1 \\
\addlinespace
E4B base & 32 & 30 & 68.7 & 1713 & 121~s & 2026.5$^\dagger$ \\
E4B QLoRA & \textbf{47} & \textbf{47} & \textbf{84.7} & 1283 & 106~s & 2026.4.1 \\
\bottomrule
\end{tabular}

\caption{Results on the held-out variants (\NumSubtasks{} subtasks $\times$ 2 variants $\times$
\NumSeeds{} seeds).
A run passes when it scores 100.
Bold: the highest pass count and mean score within each model (ties are all bold).
$^*$Without \HeldOutSeen, whose prompt equals that of a training example.
Time per run is the wall time of the answer, thinking included; it depends on the OpenVINO version
and is only indicative.
$^\dagger$2026.5 nightly build; for E2B base only the runs with seeds 5001--5002 on four subtasks.}
\label{tab:main}
\end{table}

\paragraph{Paired comparison.}
Two models answered the same 66 prompts with the same seeds, so we compare them run by run and count
the runs that only one of them passes.
The seed pairs the runs only loosely, since the sampled answer also depends on the adapter and on
small numerical differences, so the pairing mainly matches prompts.
We use the exact two-sided sign test (McNemar's test) on the discordant pairs.
For E2B, \EtwoBaseEtwoLoraOnlyB{} runs pass only with LoRA and \EtwoBaseEtwoLoraOnlyA{} only without it
($p = \EtwoBaseEtwoLoraP$); with QLoRA the counts are \EtwoBaseEtwoQloraOnlyB{} and
\EtwoBaseEtwoQloraOnlyA{} ($p = \EtwoBaseEtwoQloraP$).
For E4B, \EfourBaseEfourQloraOnlyB{} runs pass only with the adapter and \EfourBaseEfourQloraOnlyA{}
only without it ($p = \EfourBaseEfourQloraP$).
LoRA and QLoRA on E2B are indistinguishable: \EtwoLoraEtwoQloraOnlyA{} runs pass only with one of them
and \EtwoLoraEtwoQloraOnlyB{} only with the other.
The adapted E2B passes more runs than the unadapted E4B (\EtwoLoraPass{} and \EtwoQloraPass{} against
\EfourBasePass), but the difference is not significant (\EfourBaseEtwoLoraOnlyB{} against
\EfourBaseEtwoLoraOnlyA{} runs, $p = \EfourBaseEtwoLoraP$, for LoRA).
With 66 runs per model, only large differences reach significance.

\begin{figure}[t]
\centering
\includegraphics[width=\textwidth]{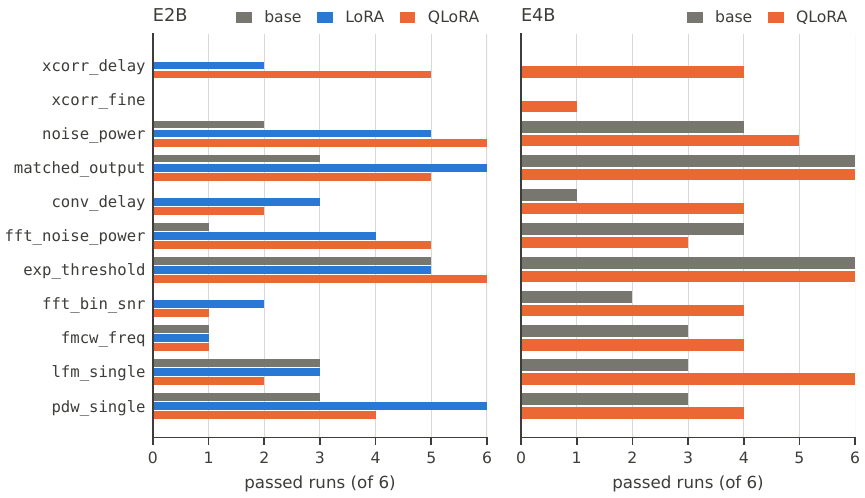}
\caption{Passed runs per subtask (of 6: two held-out variants, three seeds) for E2B without an
adapter and with the LoRA and QLoRA adapters (left), and for E4B without and with the QLoRA adapter
(right).
A missing bar means no run passed.}
\label{fig:subtasks}
\end{figure}

\paragraph{Subtasks.}
\Figref{fig:subtasks} breaks the passes down by subtask.
The largest gains are on subtasks where the base models broke conventions of MATLAB and DSP, such
as the sign and offset of a lag (\secref{sec:errors} classifies the errors).
The integer delay from \code{xcorr} goes from \EtwoBasePassXcorrGecikme{} to \EtwoLoraPassXcorrGecikme{}
(LoRA) and \EtwoQloraPassXcorrGecikme{} (QLoRA) passes for E2B and from \EfourBasePassXcorrGecikme{} to
\EfourQloraPassXcorrGecikme{} for E4B, and the delay from the matched-filter peak from
\EtwoBasePassKonvGecikme{} to \EtwoLoraPassKonvGecikme{} and \EtwoQloraPassKonvGecikme{} for E2B and from
\EfourBasePassKonvGecikme{} to \EfourQloraPassKonvGecikme{} for E4B.
The fractional delay in noise remains the hardest subtask: one run passes, with the E4B adapter.
For E2B, the FMCW frequency profile (\EtwoBasePassFmcwFrekans{}, \EtwoLoraPassFmcwFrekans{} and
\EtwoQloraPassFmcwFrekans{} passes) and the single chirp (\EtwoBasePassLfmTek{}, \EtwoLoraPassLfmTek{} and
\EtwoQloraPassLfmTek{}) do not improve; for E4B the chirp goes from \EfourBasePassLfmTek{} to
\EfourQloraPassLfmTek.
On the theoretical FFT noise power E4B gets slightly worse with the adapter
(\EfourBasePassFftGurultuGucu{} to \EfourQloraPassFftGurultuGucu{} passes), while E2B improves
(\EtwoBasePassFftGurultuGucu{} to \EtwoLoraPassFftGurultuGucu{} and \EtwoQloraPassFftGurultuGucu).

\paragraph{OpenVINO versions.}
The E4B base model was evaluated with the 2026.5 nightly build and its adapter with 2026.4.1
(\secref{sec:training}), so the E4B comparison mixes two versions.
We did not repeat the 66 adapter runs on the nightly build, for three reasons.
Both versions run the same int8 weights with the same template and sampling, and fixed seeds do not
reproduce an answer exactly even within one version, so a version change acts like a change of
seed.
With the adapter, both versions generate at the same speed (\secref{sec:training}).
And the one adapter run we repeated on the nightly build, \CheckRunTask, scored \CheckRunNightly{} as it
did on 2026.4.1.
For E2B base, \EtwoBaseNightlyPass{} of the \EtwoBaseNightlyRuns{} nightly runs on four subtasks passed,
and \EtwoBaseOldPassFour{} of the \EtwoBaseOldRunsFour{} runs of the same subtasks on 2026.4.1.
A version effect of the size of the differences in \tabref{tab:main} would be needed to change the
conclusions; we have no sign of one, but we cannot exclude a smaller one.

\section{Error analysis}\label{sec:errors}

\begin{figure}[t]
\centering
\includegraphics[width=\textwidth]{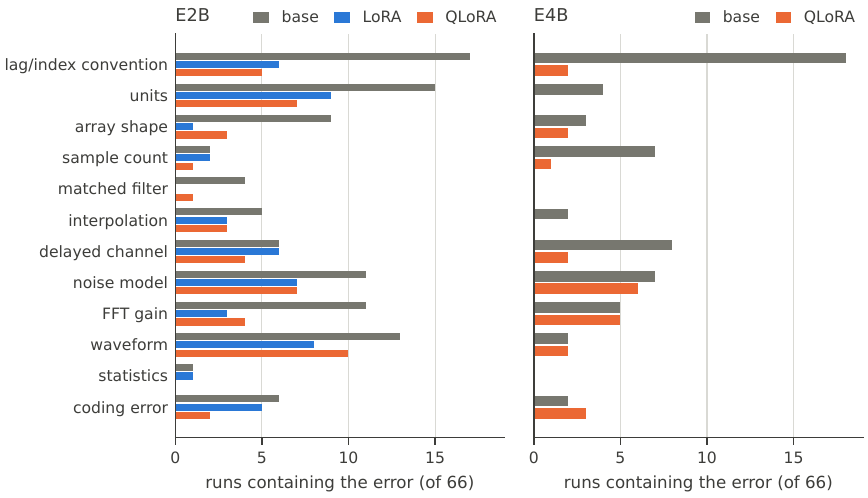}
\caption{Runs (of 66) whose code contains each error class, for E2B without an adapter and with the
LoRA and QLoRA adapters (left) and for E4B without and with the QLoRA adapter (right).
Only failing runs were classified; a run can contain several classes.}
\label{fig:errors}
\end{figure}

The checkers tell which value is wrong, not why.
To see what the adapters change, every failing code of the five models, \NumFailsRead{} in all, was
read together with its output and given one or more of \NumErrClasses{} error classes.
The reading and labelling were done with Claude Opus 5.5.
A run can contain several errors; we count, for each model, the runs out of 66 whose code contains
a class (\figref{fig:errors}).
\Appref{app:errors} defines the classes and gives the counts as a table.
Two limits apply.
Only failing runs were read, so an error that does not change a printed value is not counted, and
when a code stops with an error the later lines may hide further errors.

\paragraph{Conventions are learned.}
The classes that shrink most are conventions that a reference solution states once and the model
can copy.
Lag and index conventions fall from \EtwoBaseErrConvention{} runs to \EtwoLoraErrConvention{} (LoRA)
and \EtwoQloraErrConvention{} (QLoRA) for E2B and from \EfourBaseErrConvention{} to
\EfourQloraErrConvention{} for E4B.
Without an adapter both models often call \code{xcorr(s1, s2)}, whose peak lies at the negative
lag, and print the delay with the wrong sign; they read the \code{conv} peak index as the delay; and
they add one to the 1-based index returned by \code{max}.
Unit errors fall from \EtwoBaseErrUnits{} to \EtwoLoraErrUnits{} and \EtwoQloraErrUnits{} for E2B and
from \EfourBaseErrUnits{} to \EfourQloraErrUnits{} for E4B, row--column errors from
\EtwoBaseErrShape{} to \EtwoLoraErrShape{} and \EtwoQloraErrShape{} for E2B, and the matched filter
without time reversal from \EtwoBaseErrMatched{} to \EtwoLoraErrMatched{} and \EtwoQloraErrMatched.
E2B still makes unit errors in \EtwoLoraErrUnits{} (LoRA) and \EtwoQloraErrUnits{} (QLoRA) runs, mostly
by writing the chirp rate of the delay subtasks with the wrong power of ten (for example
\code{k = 1.808e6} for 1.808~MHz/\textmu s, which is $1.808\times10^{12}$~Hz/s), although every
training example of these subtasks converts it.

\paragraph{Signal models are not.}
Errors in how a signal is built change much less.
For E2B, waveform errors fall only from \EtwoBaseErrWaveform{} to \EtwoLoraErrWaveform{} and
\EtwoQloraErrWaveform{}, and \EtwoChirpWaveform{} of the \EtwoChirpFails{} failures of the chirp and
FMCW subtasks in the three E2B models build the chirp phase or the frequency profile wrong:
the phase written as $2\pi f(t)\,t$ instead of the integral of $f(t)$, the slope divided twice by
the ramp duration, one ramp instead of four.
Building the delayed channel as $s(t-\tau)$ also stays hard for E2B (\EtwoBaseErrDelayed{},
\EtwoLoraErrDelayed{} and \EtwoQloraErrDelayed{} runs): the models shift the window to
$t\ge\tau$ but evaluate the phase at $t-\tau$, or shift by a whole number of samples.
The noise model is wrong in \EtwoBaseErrNoise, \EtwoLoraErrNoise{} and \EtwoQloraErrNoise{} E2B runs
and in \EfourBaseErrNoise{} and \EfourQloraErrNoise{} E4B runs.
For the E4B adapter it is the main remaining error: in \EfourQloraXinceNoiseOnly{} of its
\EfourQloraXinceFails{} failures on the fractional delay the printed delay is right and only the noise
level is wrong, because the pulse power is averaged over the whole vector, zeros included.
No training example does this; all of them pass the noise-level check.

\paragraph{Knowledge of the FFT gain.}
The theoretical value of an unscaled FFT needs the factor $N$: the mean of $|X|^2$ is $N\sigma^2$
and the SNR gain is $10\log_{10}N$.
Without an adapter both models leave it out in most failing runs of these subtasks.
For E2B the adapters reduce this error from \EtwoBaseErrFFTgain{} to \EtwoLoraErrFFTgain{} and
\EtwoQloraErrFFTgain{} runs; for E4B it stays at \EfourBaseErrFFTgain{} and \EfourQloraErrFFTgain.
Six training examples per subtask were not enough to change this answer of E4B.

\paragraph{Examples.}
The following lines are taken from failing answers, shortened and with variable names abbreviated;
the comments are ours.
\begin{lstlisting}[basicstyle=\ttfamily\footnotesize]
[R, lags] = xcorr(s1, s2);         % E4B base: prints -31, not 31
s2 = circshift(s, round(tau*fs));  % E4B base: delay 9.31 becomes 9
P = mean(abs(channel1).^2);        % E4B QLoRA: zeros outside the pulse lower P
theory = variance_x;               % E4B QLoRA: N*variance_x expected
k = 11.595e6;                      % E2B LoRA: 11.595 MHz/us = 11.595e12 Hz/s
n = sigma*randn(N, 1);             % E4B base: column noise + row tone: N x N
chirp = exp(1i*2*pi*f.*t);         % E2B base: the phase must integrate f
h = conj(s); y = conv(s, h);       % E2B base: needs conj(fliplr(s))
\end{lstlisting}

\section{Limitations}\label{sec:limitations}

\paragraph{Emulator instead of MATLAB.}
All code runs in our own emulator of a MATLAB subset (\secref{sec:emulator}), not in MATLAB.
A difference between the two can fail a correct answer or pass a wrong one.
We know of one probable case: an E4B adapter answer on \TaskXcorrDelay{} assigns into a scalar with
a logical mask longer than the scalar.
MATLAB grows the array in this case, as far as we know, while the emulator stops with an error.
We could not check this in MATLAB, and the answer would not necessarily have passed there.

\paragraph{Checkers.}
The checkers test the printed values and some arrays of the workspace, not every line of the code.
Reading the training examples showed that they had accepted wrong code (\secref{sec:data}).
The noise-level check of \TaskXcorrFine{} was added before the final scoring, after a passing
evaluation answer turned out to set the noise from the power of the whole vector; all evaluations
were scored again with the final checkers.
Other errors that do not change a checked value can remain in passing runs; for example, no check
tests whether the two channels of \TaskXcorrFine{} get independent noise.

\paragraph{Small samples.}
Each model was evaluated on 66 runs, six per subtask, and each adapter comes from one training run
with settings we did not tune.
Differences of a few passes between two models, or of one or two passes within a subtask, are
within the noise.
The seeds pair two models' runs only loosely, so the paired tests mainly match prompts.

\paragraph{Training data.}
The \NumExamples{} examples were made with expert reference solutions, filters, repeated sampling and
hand fixes.
\NumHandFixed{} examples were fixed by hand, \NumCodeFixed{} of them in what the code computes, and
\NumRulesStepThree{} answers were obtained with a system message whose facts were written by a larger
model (\secref{sec:data}).
The models therefore learn from text that E2B wrote under guidance, not from E2B's unaided output.
The training data are not released.

\paragraph{What the evaluation covers.}
The held-out variants come from the same templates as the training variants, and one of them,
\HeldOutSeen, even has the same prompt as a training example (results without it are in
\tabref{tab:main}).
The evaluation thus measures new parameters of known subtasks.
We did not test new subtasks, longer programs or whether the adapters harm the models on other
tasks.

\paragraph{Software and hardware.}
Two OpenVINO versions were used (\secref{sec:results}), the adapters were trained on bf16 or NF4
weights and run on int8 weights, and all measurements come from one laptop.
Speeds and run times depend on the software version and on system settings and may differ on other
machines.

\paragraph{Error labels.}
The error classes of \secref{sec:errors} were assigned by reading the failing codes with Claude
Opus 5.5, and only failing runs were read.
Another reader could draw some class boundaries differently, for example between a unit error and a
waveform error caused by it.

\section{Conclusion}\label{sec:conclusion}

Sixty-six checked examples, one per training variant, were enough to change how two small models
write MATLAB signal-processing code.
With a LoRA or QLoRA adapter trained in under an hour on the laptop's integrated GPU, E2B passed
\EtwoLoraPass{} and \EtwoQloraPass{} of 66 held-out runs instead of \EtwoBasePass, and E4B
\EfourQloraPass{} instead of \EfourBasePass; QLoRA was as good as LoRA on E2B.
The error analysis shows what such a small set teaches: conventions that a solution states once,
such as the sign of a lag, the unit of a chirp rate or the shape of an array, largely disappear.
Errors in building a signal, such as the phase of a chirp, a delayed channel or the level of the
noise, and the FFT gain for E4B, change little.
These need more than six examples per subtask, or more examples aimed at the signal model.

The adapter costs \EtwoTpsDropPct--\EfourTpsDropPct\% of the generation speed when it is applied at run
time; merging it into the int8 weights would remove this cost.
Further work should test the adapters on new subtasks and on tasks outside signal processing, and
check the emulator against MATLAB.

\vskip1cm
\noindent {\it Acknowledgements}
The numbered facts of the generation-time system message (\secref{sec:data}) were written by
Claude Opus 5.5.
Assistance on many details was provided by Claude Opus 5.5.

\bibliographystyle{unsrtnat}
\bibliography{refs}

\appendix
\section{Task prompts}\label{app:prompts}

This appendix gives the prompt of held-out variant 6 of every subtask, as the models received it,
and the values the checker expects for that variant.
Each prompt was sent after the system message of \secref{sec:protocol}.
The prompts, the expected values and the reference solutions are generated by the task templates;
the training data are not released.
In the prompt of \code{pdw\_single}, RF is the carrier frequency, PW the pulse width, PA the pulse
amplitude and DOA the direction of arrival.

\paragraph{1. \code{xcorr\_delay}: integer delay between two LFM channels from the \code{xcorr} peak.}
\begin{quote}\small\ttfamily\raggedright
Two channels are sampled at fs = 20 MHz, 1024 samples each (t = (0:N-1)/fs). Channel 1 contains a complex LFM pulse s(t) = exp(1j*pi*k*(t - tc).\textasciicircum{}2) for t0 <= t < t0 + Tp and zero elsewhere, with t0 = 12750 ns, Tp = 2200 ns, k = 3.17 MHz/us and tc = t0 + Tp/2. Channel 2 contains s(t - tau) with tau = 1545 ns. There is no noise. Estimate the delay of channel 2 relative to channel 1 in whole samples from the peak of the cross-correlation computed with xcorr and print it.
\end{quote}
Expected values: integer delay (from \code{xcorr}) $31$.

\paragraph{2. \code{xcorr\_fine}: fractional delay in noise: \code{xcorr} peak refined by a parabola.}
\begin{quote}\small\ttfamily\raggedright
Two channels are sampled at fs = 20 MHz, 256 samples each (t = (0:N-1)/fs). Channel 1 contains a complex LFM pulse s(t) = exp(1j*pi*k*(t - tc).\textasciicircum{}2) for t0 <= t < t0 + Tp and zero elsewhere, with t0 = 2250 ns, Tp = 3500 ns, k = 1.808 MHz/us and tc = t0 + Tp/2. Channel 2 contains s(t - tau) with tau = 465.5 ns. Add independent complex white Gaussian noise to each channel with an SNR of 30 dB relative to the pulse power. Estimate the delay of channel 2 relative to channel 1 in samples: find the peak of the cross-correlation magnitude computed with xcorr, then refine it with parabolic interpolation through the peak and its two neighbours. Print the refined delay in samples with 4 decimals.
\end{quote}
Expected values: refined delay $9.31 \pm 0.2$.

\paragraph{3. \code{noise\_power}: complex tone plus noise at a given SNR; measured powers.}
\begin{quote}\small\ttfamily\raggedright
Generate 16384 samples of a complex tone 2*exp(1j*2*pi*f0*t) with f0 = 0.74 MHz and fs = 10 MHz, and add complex white Gaussian noise so that the SNR (tone power to noise power) is 15 dB; call the noisy signal x. From the noise you generated, print the measured noise power (mean of |noise|\textasciicircum{}2) and the measured variance of its real part, and print the measured power of x (mean of |x|\textasciicircum{}2).
\end{quote}
Expected values: noise power $0.12649 \pm 0.013$; variance of the real part $0.063246 \pm 0.0063$; power of $x$ $4.1265 \pm 0.04$.

\paragraph{4. \code{matched\_output}: matched filter of an LFM pulse with \code{conv}: length, peak index, peak.}
\begin{quote}\small\ttfamily\raggedright
Generate a complex baseband LFM pulse with amplitude 3, bandwidth 2 MHz and duration 10 us, sampled at 5 MHz (N = T*fs samples, t = (0:N-1)/fs - T/2, s = A*exp(1j*pi*(B/T)*t.\textasciicircum{}2)). Apply its matched filter to the pulse with conv and print the length of the filter output, the 1-based index of its peak magnitude and the peak magnitude.
\end{quote}
Expected values: length $99$; peak index $50$; peak magnitude $450 \pm 0.45$.

\paragraph{5. \code{conv\_delay}: delay of a pulse in a receive window from the matched-filter peak.}
\begin{quote}\small\ttfamily\raggedright
Generate a complex baseband LFM pulse with amplitude 0.8, bandwidth 2 MHz and duration 40 us, sampled at 10 MHz (N = T*fs samples, t = (0:N-1)/fs - T/2, s = A*exp(1j*pi*(B/T)*t.\textasciicircum{}2)). A receiver records 504 samples with no noise: zeros except this pulse, which starts 4.6 us after the first sample. Apply the matched filter with conv(rx, h), where h is the matched filter of the pulse, and print the 1-based index of the output peak and the delay of the pulse in samples obtained from that index.
\end{quote}
Expected values: peak index $446$; delay from the peak index $46$.

\paragraph{6. \code{fft\_noise\_power}: mean $|X|^2$ of complex noise after an unscaled FFT, measured and theoretical.}
\begin{quote}\small\ttfamily\raggedright
Generate 1024 samples of complex white Gaussian noise with variance 4 (total power of each complex sample) and compute X = fft(x) without any scaling. Print the measured mean of |X|.\textasciicircum{}2 over all bins and its theoretical value.
\end{quote}
Expected values: measured mean $4096 \pm 410$; theoretical mean $4096 \pm 4.1$.

\paragraph{7. \code{exp\_threshold}: threshold on $|X(k)|^2$ for a false-alarm probability (exponential law).}
\begin{quote}\small\ttfamily\raggedright
In an FFT of pure complex white Gaussian noise, |X(k)|\textasciicircum{}2 in each bin is exponentially distributed with mean 8192. Compute the threshold T on |X(k)|\textasciicircum{}2 for which the probability of exceeding it is 0.01, and print T and the corresponding threshold on |X(k)|.
\end{quote}
Expected values: threshold on $|X|^2$ $37726 \pm 75$; threshold on $|X|$ $194.23 \pm 0.39$.

\paragraph{8. \code{fft\_bin\_snr}: tone in noise: peak bin, measured and theoretical SNR after the FFT.}
\begin{quote}\small\ttfamily\raggedright
A complex tone 0.5*exp(1j*2*pi*f0*t) with f0 = 70373.5 Hz is in complex white Gaussian noise with an SNR of -10 dB per sample; fs = 1 MHz and 16384 samples, t = (0:N-1)/fs. Compute X = fft(x) and print the 1-based index of the largest |X|, the measured SNR at the FFT output in dB (|X|\textasciicircum{}2 at that bin divided by the mean |X|\textasciicircum{}2 of all other bins) and the theoretical SNR at the FFT output in dB.
\end{quote}
Expected values: peak bin (1-based) $1154$; measured SNR $32.144 \pm 1.5$; theoretical SNR $32.144 \pm 0.02$.

\paragraph{9. \code{fmcw\_freq}: instantaneous-frequency vector of a sawtooth or triangular FMCW signal.}
\begin{quote}\small\ttfamily\raggedright
An FMCW signal uses sawtooth down-chirp modulation sweeping from +B/2 down to -B/2 with B = 50 MHz, ramp duration 40 us, sampled at 75 MHz, with 4 ramps. Build the vector of its instantaneous frequency in MHz at every sample (sample n at time (n-1)/fs) and print its length, its minimum, its maximum and its value at t = 39.8267 us.
\end{quote}
Expected values: length $12000$; minimum $-25 \pm 0.025$; maximum $25 \pm 0.025$; value at a given time $-24.783 \pm 0.0083$.

\paragraph{10. \code{lfm\_single}: one complex baseband linear chirp stored in a vector.}
\begin{quote}\small\ttfamily\raggedright
Write MATLAB code that generates a single complex baseband linear chirp sweeping from +B/2 to -B/2 with B = 10 MHz over 40 us, sampled at 20 MHz. Store it in a single vector and print its length.
\end{quote}
No printed value is checked; the stored vector is checked in the workspace.

\paragraph{11. \code{pdw\_single}: pulse descriptor words (PDW matrix) of one emitter with constant pulse repetition interval (PRI).}
\begin{quote}\small\ttfamily\raggedright
Write a PDW generator for one emitter. A PDW list is an N x 5 matrix with columns [TOA(us) RF(MHz) PW(us) PA(dBm) DOA(deg)], one row per pulse. The emitter has PRI 80 us, RF 9364 MHz, PW 1 us, PA -31 dBm, DOA 292 deg and its first pulse at TOA 63.5 us; constant PRI, no jitter. Build the PDW matrix of every pulse with TOA < 2000 us and print the number of PDWs and the TOA of the last one.
\end{quote}
Expected values: number of PDWs $25$; last TOA $1983.5 \pm 0.001$.

\section{Error classes}\label{app:errors}

\Tabref{tab:errors} defines the error classes of \secref{sec:errors} and gives, for each model, the
number of runs (of 66) whose code contains the class.
A run can contain several classes, so a column does not sum to the number of failing runs.

\begin{table}[h]
\centering\small
\begin{tabular}{@{}>{\raggedright\arraybackslash}p{0.15\textwidth}>{\raggedright\arraybackslash}p{0.37\textwidth}rrrrr@{}}
\toprule
& & \multicolumn{3}{c}{E2B} & \multicolumn{2}{c}{E4B} \\
\cmidrule(lr){3-5}\cmidrule(l){6-7}
Class & Definition & base & LoRA & QLoRA & base & QLoRA \\
\midrule
lag/index convention & lag sign from the \code{xcorr} argument order, zero-lag index, delay read from the \code{conv} peak without the pulse-length offset, 0-based index of \code{max} & 17 & 6 & 5 & 18 & 2 \\
units & MHz/\textmu s not converted to Hz/s, \textmu s mixed with s & 15 & 9 & 7 & 4 & 0 \\
array shape & row and column vectors mixed (an $N\times N$ matrix), wrong matrix layout & 9 & 1 & 3 & 3 & 2 \\
sample count & number of samples or pulses or the time grid off by one (\code{ceil(T*fs)}, \code{linspace(0,T,N)}, a missing $+1$) & 2 & 2 & 1 & 7 & 1 \\
matched filter & matched filter without time reversal (\code{conj(s)} or \code{s}) & 4 & 0 & 1 & 0 & 0 \\
interpolation & parabolic interpolation wrong: formula, sign, fixed coefficient, rounding, index instead of lag & 5 & 3 & 3 & 2 & 0 \\
delayed channel & channel 2 not built as $s(t-\tau)$: advanced, shifted by a whole number of samples, or window shifted while the phase is not & 6 & 6 & 4 & 8 & 2 \\
noise model & noise level or independence wrong: pulse power averaged over the whole vector, missing factor 1/2, one noise vector for both channels, energy instead of power, inverted SNR & 11 & 7 & 7 & 7 & 6 \\
FFT gain & gain $N$ of the unscaled FFT missing in a theoretical value ($N\sigma^2$, $10\log_{10}N$) & 11 & 3 & 4 & 5 & 5 \\
waveform & chirp phase, pulse window or FMCW frequency profile wrong & 13 & 8 & 10 & 2 & 2 \\
statistics & exponential threshold with $\ln(1-p)$ or with the mean taken as the rate & 1 & 1 & 0 & 0 & 0 \\
coding error & code that stops with an error: undefined variable, infinite range, fractional index, syntax error & 6 & 5 & 2 & 2 & 3 \\
\midrule
failing runs & & 48 & 29 & 29 & 34 & 19 \\
\bottomrule
\end{tabular}

\caption{Error classes and the number of runs (of 66) whose code contains each class.
Only failing runs were classified.}
\label{tab:errors}
\end{table}

\end{document}